\documentclass[letterpaper]{article} % DO NOT CHANGE THIS
\usepackage[preprint]{aaai2027}  % Use [submission] for anonymous AAAI review
\usepackage[hyphens]{url} % DO NOT CHANGE THIS
\usepackage{graphicx} % DO NOT CHANGE THIS
\usepackage{natbib}  % DO NOT CHANGE THIS
\usepackage{caption} % DO NOT CHANGE THIS
\usepackage{amsmath}
\usepackage{amssymb}
\usepackage{amsfonts}
\usepackage{algorithm}
\usepackage{algorithmic}
\usepackage{booktabs}
\usepackage{multirow}

\title{UniTexture: Cross-Task Universal Adversarial Textures for Vision-Language-Action Models}

\author{
Yukun Dai,
Mingzhe Dai,
Tianshi Wang\textsuperscript{\rm 1},
Fengling Li\textsuperscript{\rm 2},\\
Jingjing Li\textsuperscript{\rm 3},
Lei Zhu\textsuperscript{\rm 1}
}
\affiliations{
\textsuperscript{\rm 1}Tongji University\\
\textsuperscript{\rm 2}Mohamed bin Zayed University of Artificial Intelligence\\
\textsuperscript{\rm 3}University of Electronic Science and Technology of China\\
yukundai@tongji.edu.cn, daimingzhe@bupt.edu.cn,\\
tswang0116@163.com, fenglingli2023@gmail.com,\\
lijin117@yeah.net, leizhu0608@gmail.com
}

\begin{document}

\maketitle

\begin{abstract}
Vision-Language-Action (VLA) models have emerged as generalist robotic policies capable of following diverse language instructions and performing a wide range of manipulation tasks. However, their direct control over embodied agents also exposes them to adversarial interference that may cause unsafe physical behaviors. Existing attacks on robotic policies are typically optimized for a single task or instruction, leaving the cross-task vulnerabilities of multitask VLAs largely unexplored. We introduce UniTexture, a cross-task universal adversarial texture attack that uses a single textured 3D object to induce targeted deviations in VLA action predictions across multiple tasks. UniTexture backpropagates gradients from the policy’s action outputs to surface texture parameters through a differentiable renderer. It jointly optimizes the shared texture over a distribution of tasks, instructions, states, and viewpoints using a targeted action-space objective, steering predicted actions toward attacker-defined targets without optimizing a separate texture for each task. We evaluate UniTexture on OpenVLA and $\pi_{0.5}$ across diverse manipulation tasks and multiple evaluation settings. UniTexture reduces the mean task success rate from $90.0\%$ under benign conditions to $48.4\%$ under attack, induces target-aligned action shifts, and further exhibits cross-suite and cross-model transfer without re-optimization. Together, these findings reveal shared cross-task vulnerabilities in multitask VLAs that can be systematically exploited through a single adversarial surface texture.
\end{abstract}

\section{Introduction}

Vision-language-action (VLA) models are emerging as generalist policies for robotic manipulation. By grounding natural-language instructions and visual observations in robot actions, a single policy can execute diverse tasks without requiring a separately engineered controller for each behavior~\citep{brohan2022rt,pmlr-v229-zitkovich23a,o2024open,kim2024openvla,black2024pi_0}. This versatility, however, also broadens the attack surface: a single adversarial visual pattern may influence multiple behaviors governed by the same policy. Moreover, an erroneous VLA prediction is not merely a semantic mistake. Because the predicted action is executed in the environment, adversarial interference may lead to unsafe motion, failed manipulation, or physical damage.

\begin{figure}[!t]
\centering
\includegraphics[pagebox=cropbox,width=\columnwidth]{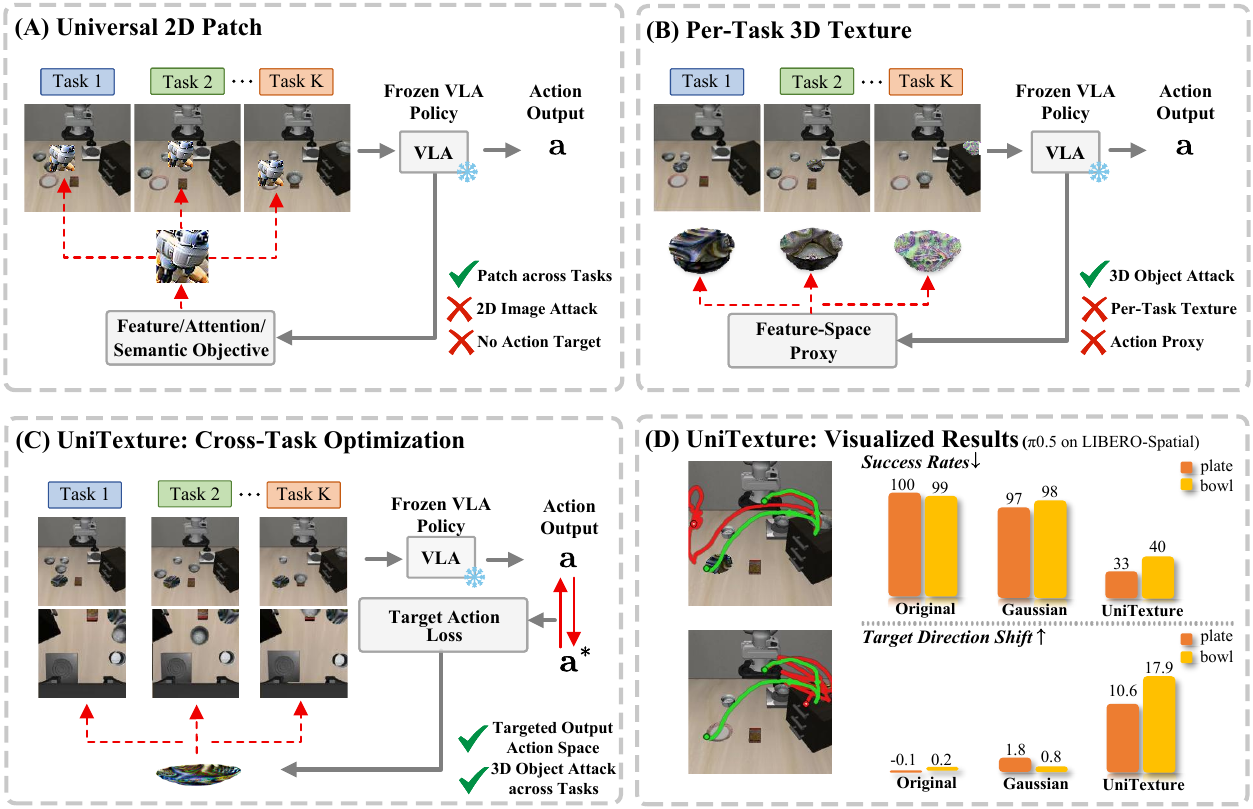}
\caption{Overview and comparison of UniTexture. (A) Universal 2D patches transfer across tasks but lack 3D surface constraints and explicit action targets. (B) Existing 3D texture attacks are object-bound but require task-specific textures and feature-space proxies. (C) UniTexture optimizes one shared object-bound texture across tasks using targets in the VLA’s native action space. (D) On $\pi_{0.5}$ in LIBERO-Spatial, UniTexture reduces task success and induces larger target-direction action shifts than the original and Gaussian baselines.}
\label{fig:cover}
\end{figure}

Existing visual attacks on VLAs have progressed along several complementary directions. Image-space attacks demonstrate the sensitivity of VLA action predictions to adversarial observations~\citep{wang2024exploring}. Universal physical patches further show that a single two-dimensional pattern can affect multiple tasks, viewpoints, and model architectures~\citep{lu2026robotsobeypatchuniversal}. Separately, Tex3D maps adversarial textures onto three-dimensional object surfaces, preserving their spatial correspondence as the object moves across viewpoints and interaction trajectories~\citep{chen2026tex3d}. However, universal patches lack object-surface constraints and direct targeted control in the policy’s action space, whereas Tex3D optimizes separate textures for individual tasks. Whether a single geometry-aware texture can consistently control a multitask VLA without per-task optimization therefore remains unclear.

This setting is more demanding than causing an isolated task failure. Across tasks, the instruction, scene context, nominal action distribution, and interaction trajectory all change. The target object's pose, visibility, and image footprint also evolve during manipulation and across camera views. Consequently, gradients from different tasks must produce one texture that remains effective under these coupled variations. Targeted control introduces a further requirement: the objective must encode the attacker's intended behavior in the policy's action space, rather than treating arbitrary feature displacement as a proxy for success. The attack must satisfy these requirements while remaining fixed on the same physical object.

We introduce \textbf{UniTexture}, a cross-task universal adversarial texture attack for VLAs, as illustrated in Figure~\ref{fig:cover}. Given a frozen target VLA, a target object, and a collection of manipulation tasks, UniTexture jointly optimizes a single shared surface texture over a cross-task training distribution. We first calibrate a differentiable renderer using clean multi-view observations by aligning the scene geometry and estimating photometric parameters, after which all rendering parameters are frozen. During attack optimization, we use task-balanced sampling to obtain task-conditioned observations, object poses, and language instructions. The renderer composites the shared texture into each observation, while targeted objectives defined in the model’s native action representation propagate gradients from the VLA outputs to the texture parameters. Only the texture is updated, and the resulting texture is applied unchanged across all tasks, without separate per-task optimization or refinement.

We evaluate UniTexture on OpenVLA~\citep{kim2024openvla} and $\pi_{0.5}$~\citep{intelligence2025pi05visionlanguageactionmodelopenworld} using LIBERO-Spatial and LIBERO-Goal~\citep{liu2023libero}. For each model-object-suite configuration, a single texture is jointly optimized for all tasks in the corresponding suite. Across the evaluated settings, UniTexture reduces the mean task success rate from $90.0\%$ under benign conditions to $48.4\%$ under attack, while consistently shifting predicted actions toward attacker-specified targets. Cross-suite and cross-model evaluations further demonstrate that the optimized textures retain attack effectiveness beyond the task distribution and model used during optimization, without re-optimization. These findings expose a deployment-level vulnerability: task diversity alone does not prevent a single adversarial object appearance from repeatedly influencing a multitask robotic policy.

Our contributions are summarized as follows:
\begin{itemize}
    \item \textbf{Cross-task universal 3D texture attack.} We formulate a threat model in which a single object-bound texture is jointly optimized across multiple tasks and applied unchanged, without per-task optimization or refinement.
    \item \textbf{Targeted action-space optimization.} We develop model-compatible objectives and metrics that directly encode attacker-specified targets in each VLA’s native action space, rather than relying on feature-space proxies or untargeted failure.
    \item \textbf{Comprehensive cross-task evaluation.} Experiments on OpenVLA and $\pi_{0.5}$ across LIBERO-Spatial and LIBERO-Goal demonstrate cross-task attack effectiveness, targeted action manipulation, and transfer across task suites and model architectures.
\end{itemize}

\section{Related Work}

\subsection{Vision-Language-Action Models}
Vision-language-action (VLA) models integrate visual perception, language conditioning, and action generation into unified policies for robot control~\citep{pmlr-v229-zitkovich23a}. Because a single policy is reused across diverse tasks, shared visual inputs create a common attack surface across behaviors. However, VLA architectures differ substantially in their action representations: OpenVLA~\citep{kim2024openvla} uses autoregressive action tokens, the $\pi$ family~\citep{black2024pi_0,intelligence2025pi05visionlanguageactionmodelopenworld} employs flow-matching action experts, and recent variants introduce continuous prediction or alternative action tokenizations~\citep{kim2025openvlaoft,pertsch2025fast}. This heterogeneity motivates attack objectives defined in each model's native action space.

\subsection{Adversarial Attacks on VLA Models}
Building on classical digital and physical attacks~\citep{szegedy2014intriguing,madry2018towards,eykholt2018robust,athalye2018synthesizing}, VLA attacks include training-time backdoors~\citep{zhou2025badvla}, language-based attacks~\citep{jones2025adversarial}, and deployment-time visual attacks~\citep{wang2024exploring,cheng2024manipulation,lu2026phantom,zhang2026redvla,guo2025robustness,cui2026liberosafety}. Existing visual attacks mainly pursue universality or geometric persistence. Image-space attacks reveal vulnerabilities in VLA action prediction~\citep{wang2024exploring}, while UPA-RFAS~\citep{lu2026robotsobeypatchuniversal} and VLA-Hijack~\citep{fu2026vlahijack} optimize transferable 2D patches through feature-, attention-, or semantic-level objectives. These patches, however, are composited in the image plane rather than bound to object geometry. Tex3D~\citep{chen2026tex3d}, building on texture attacks against navigation agents~\citep{liu2020spatiotemporal}, maps adversarial textures onto 3D object surfaces through differentiable rendering, but optimizes separate textures for individual tasks and relies on a visual-feature proxy for $\pi$-family policies. UniTexture addresses the intersection of these directions by jointly optimizing one object-bound texture across tasks using attacker-specified targets in each VLA's native action space.

\section{Method}

\begin{figure*}[t]
\centering
\includegraphics[width=\textwidth]{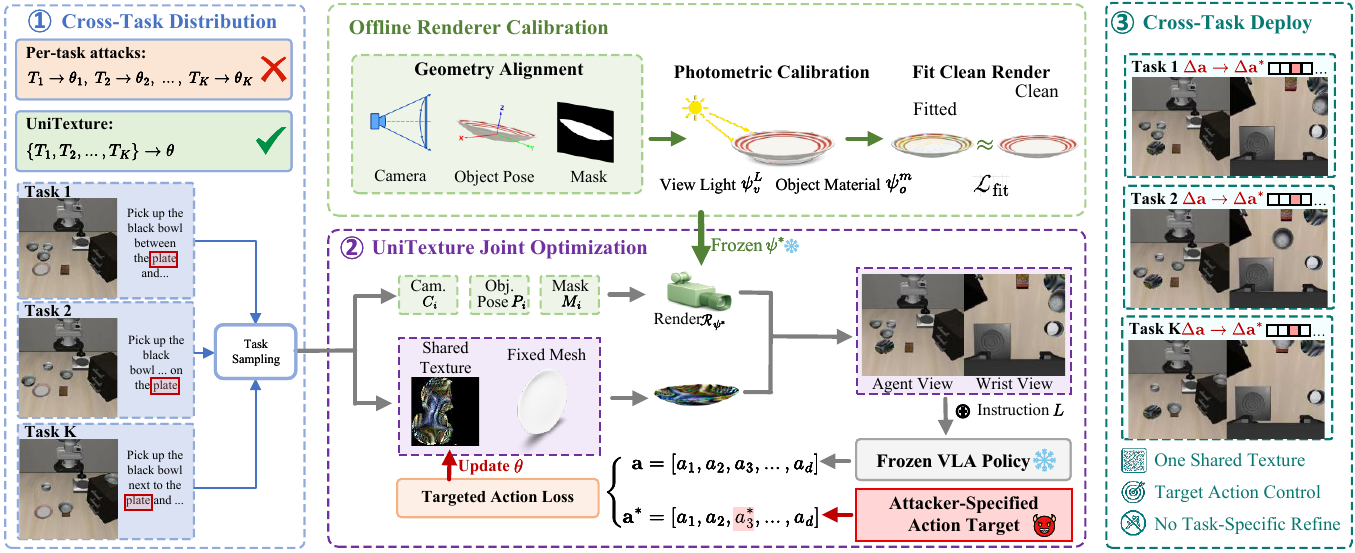}
\caption{Overview of UniTexture. \textbf{(1)} A task suite induces a cross-task distribution: rather than fitting a separate texture per task, UniTexture optimizes one shared texture $\theta$ for all tasks. Before joint optimization, an offline calibration stage fits per-view lighting $\psi^{L}_{v}$ and per-object material $\psi^{m}_{o}$ so that the rendered clean object matches the simulator's appearance; the resulting $\psi^{\star}$ is then frozen. \textbf{(2)} During joint optimization, each sampled frame supplies its rendering condition $(C_i, P_i, M_i)$; the shared texture is rendered on the fixed mesh, composited into the agent view (OpenVLA and $\pi_{0.5}$) and the wrist view ($\pi_{0.5}$), and passed with the instruction $L$ to the frozen VLA policy. A targeted action loss compares the predicted action $\mathbf{a}$ with the attacker-specified target $\mathbf{a}^{\star}$ and backpropagates to $\theta$ alone. \textbf{(3)} The optimized texture is deployed unchanged across all tasks.}
\label{fig:framework}
\end{figure*}

\subsection{Threat Model and Problem Formulation}
We attack a frozen VLA policy $F$ that maps an RGB observation and a language instruction to robot actions. The attacker controls only the \emph{surface appearance} of a single object in the workspace: it may repaint that object's texture, but cannot modify the policy weights, the instruction, the camera pose, the sensing pipeline, or any other object in the scene. Given a task suite $\mathcal{S}=\{\tau_1,\dots,\tau_K\}$, a target object, and an attacker-chosen target action $\mathbf{a}^\star$, the attacker optimizes one texture offline and then deploys the repainted object unchanged on every task in $\mathcal{S}$. Gradients of $F$ are available during optimization, but $F$ is never updated.

Let $\theta\in[0,1]^{H\times W\times 3}$ denote the UV texture map of the target object. We optimize $\theta$ while keeping the object's mesh geometry fixed. Each demonstration frame supplies a tuple $\xi=(I, L, C, P, M)$: a policy-aligned observation $I$, its instruction $L$, camera parameters $C$, the object's $6$-DoF pose $P$, and the object's image-space support mask $M$. Writing $\mathcal{R}_\psi$ for the differentiable rendering of the textured object under rendering parameters $\psi$, the attacked observation is the composite
\begin{equation}
\Omega(\theta;I,C,P,M,\psi)=M\odot\mathcal{R}_{\psi}(\theta;C,P)+(1-M)\odot I .
\label{eq:composite}
\end{equation}

Using the calibrated rendering parameters $\psi^\star$ defined below,
we write the attacked policy input as
$\widetilde I=\Omega(\theta;I,C,P,M,\psi^\star)$.
UniTexture optimizes a single shared texture over the target task distribution,
\begin{equation}
\theta^{\star}
=
\mathop{\mathrm{argmin}}\limits_{\theta}\;
\mathop{\mathbb{E}}\limits_{\tau\sim p(\tau)}
\mathop{\mathbb{E}}\limits_{\xi\sim\mathcal{D}_{\tau}}
\left[
\mathcal{L}_{\mathrm{tgt}}
\left(
F(\widetilde I,L);
\mathbf{a}^{\star}
\right)
\right].
\label{eq:objective}
\end{equation}
where $p(\tau)$ is the task-sampling distribution over $\mathcal{S}$, $\mathcal{D}_{\tau}$ is the demonstration distribution of task $\tau$, and $\mathbf{a}^\star$ is the attacker-specified target action.

\subsection{Calibrating and Freezing the Renderer}
Gradients reaching $\theta$ through \eqref{eq:composite} are only useful when the rendered object is photometrically consistent with the policy observations; otherwise, the attack may optimize against rendering artifacts. We therefore calibrate

\begin{equation}
\begin{aligned}
\psi
&=
\left(
\{\psi_v^{L}\}_{v\in\mathcal{V}},
\{\psi_o^{m}\}_{o\in\mathcal{O}}
\right),\\
\psi_v^{L}
&=
\left(
p_v^{W},
\{c_v^{q}\}_{q\in\mathcal{Q}}
\right),
& v&\in\mathcal{V},\\
\psi_o^{m}
&=
\left(
\{m_o^{q}\}_{q\in\mathcal{Q}},
m_o^{\mathrm{sh}}
\right),
& o&\in\mathcal{O},
\end{aligned}
\label{eq:renderer_parameters}
\end{equation}
comprising view-specific lighting parameters $\psi_v^{L}$ and object-specific material parameters $\psi_o^{m}$. Here, $\mathcal{V}=\{\mathrm{agent},\mathrm{wrist}\}$ denotes the
policy views, $\mathcal{O}$ denotes the calibrated objects, and
$\mathcal{Q}=
\{\mathrm{amb},\mathrm{diff},\mathrm{spec}\}$
indexes the ambient, diffuse, and specular components. For each view $v$, $p_v^{W}$ denotes the world-space light position
and $c_v^{q}$ denotes the RGB color or intensity of lighting component
$q$. For each object $o$, $m_o^{q}$ denotes the corresponding material
coefficient and $m_o^{\mathrm{sh}}$ denotes its shininess parameter.

The lighting parameters for each view are shared across all objects and tasks, whereas the material parameters for each object are shared across all views and tasks, with each $m_o^{q}$ broadcast across the RGB channels. The world-space light position is transformed into the object frame for each sample using its pose $P$. 

Let $\mathcal{D}_{\mathrm{cal}}$ denote the clean calibration
distribution obtained by first sampling a view--object pair and then
a calibration frame $\zeta=(I,C,P,M)$ for that pair. We obtain the
calibrated parameters from frames containing the original texture
$\theta_0$ by minimizing the masked photometric discrepancy:
\begin{equation}
\psi^{\star}
=
\mathop{\mathrm{argmin}}\limits_{\psi}\;
\mathop{\mathbb{E}}\limits_{\zeta\sim\mathcal{D}_{\mathrm{cal}}}
\left[
\mathcal{H}
\left(
M\odot\mathcal{R}_{\psi}(\theta_0;C,P),
M\odot I
\right)
\right].
\label{eq:fit}
\end{equation}
where $\mathcal{H}$ denotes the Smooth-L1 loss over masked object pixels, and $\psi^\star$ is fixed during texture optimization.

\subsection{Model-Specific Targeted Objectives}
To accommodate heterogeneous VLA action interfaces, we instantiate $\mathcal{L}_{\mathrm{tgt}}$ for autoregressive tokens and flow-matching action chunks.

%autoregression generate
\emph{Autoregressive action tokens.} Autoregressive action-token policies represent each action dimension as a discrete token and generate the resulting token sequence
autoregressively. We apply targeted token supervision~\citep{wang2024exploring} to optimize the shared object texture through differentiable rendering. Let $Q_{\mathrm{act}}$ denote the policy's action tokenizer. Given the attacked observation $\widetilde I$ defined above, we define
\begin{equation}
\begin{aligned}
\mathbf{z}^{\star}
&=Q_{\mathrm{act}}(\mathbf{a}^{\star}),\\
\mathcal{L}_{\mathrm{tgt}}^{\mathrm{tok}}(\theta)
&=
-\frac{1}{|\mathcal{J}|}
\sum_{j\in\mathcal{J}}
\log p_{F}
\left(z_j^\star\big|\tilde I,L\right),
\end{aligned}
\label{eq:openvla_target}
\end{equation}  % XXX 存疑，CEloss是这样的吗，检查一下
where $z_j^\star$ is the target token for action dimension $j$,
$p_F(z_j^\star\mid\widetilde I,L)$ denotes the conditional probability
that the frozen policy $F$ assigns to this token, and $\mathcal{J}$
contains the attacker-selected dimensions. In implementation, we mask
all non-target action-token positions so that they do not contribute
to the targeted loss.

\emph{Flow-matching action chunks.} Policies with flow-matching action experts generate continuous action chunks. Let $A\in\mathbb{R}^{H_a\times d}$ denote the clean action chunk associated with a demonstration frame, where $H_a$ is the number of action steps and $d$ is the total number of action dimensions. We construct the targeted chunk $A^\star$ by replacing only the attacker-selected action dimension $j$ across the entire action horizon:

\begin{equation}
A^\star_{h,k}=
\begin{cases}
\operatorname{clip}_{[-1,1]}(a^\star), & k=j,\\
A_{h,k}, & k\neq j,
\end{cases}
\quad h=1,\ldots,H_a .
\label{eq:target_chunk}
\end{equation}
Here, $h$ indexes the action steps, $k$ indexes the action
dimensions, and $a^\star$ is the attacker-specified target value.

The targeted chunk is then interpolated with Gaussian noise
$\epsilon$ at flow time $t$:
\begin{equation}
x_t=t\epsilon+(1-t)A^\star,
\qquad
u_t=\epsilon-A^\star .  %%u 要不要 t，研究一下
\label{eq:pi05_flow_path}
\end{equation}
Here, $\epsilon$ is a Gaussian noise chunk with the same shape as
$A^\star$, $t\in[0,1]$ is the flow-matching time, $x_t$ is the
corresponding noisy action chunk, and $u_t$ is its target velocity.

Given $x_t$, the frozen $\pi_{0.5}$ action expert predicts a
velocity field. We minimize its flow-matching residual only on the
attacker-selected action dimension $j$:
\begin{equation}
\mathcal{L}_{\mathrm{tgt}}^{\mathrm{flow}}(\theta)
=
\frac{1}{H_a}
\sum_{h=1}^{H_a}
\left[
v_F(\tilde I,L,x_t,t)_{h,j}
-
(u_t)_{h,j}
\right]^2 .
\label{eq:pi05_target}
\end{equation}
Here, $v_F$ denotes the velocity field predicted by the frozen
$\pi_{0.5}$ policy $F$. The indices $(h,j)$ select the targeted
action dimension at action step $h$; all non-target action
dimensions are excluded from the loss.

\subsection{Optimization}
We optimize only the shared texture $\theta$, keeping the policy $F$, object geometry, and calibrated renderer parameters $\psi^\star$ fixed. At each outer iteration, we sample a minibatch of frames and apply $R$ consecutive texture updates. For each update, we re-render the texture under the frame-specific conditions $(C_i,P_i,M_i)$, composite the object using \eqref{eq:composite}, evaluate the targeted objective, and update $\theta$ with AdamW. The texture is then clamped elementwise to $[0,1]$. Algorithm~\ref{alg:unitexture} summarizes the procedure.

\begin{algorithm}[t]
\caption{UniTexture: cross-task universal texture}
\label{alg:unitexture}
\begin{algorithmic}[1]
\REQUIRE frozen policy $F$; task suite $\mathcal{S}$; target-object mesh;
demonstration distributions $\{\mathcal{D}_{\tau}\}$; target specification
$(\mathbf{a}^{\star},\mathcal{J})$; outer iterations $N$; inner updates $R$
\STATE fit $\psi^{\star}$ on clean frames by \eqref{eq:fit};
freeze $\psi^{\star}$
\STATE initialize $\theta\leftarrow\theta_{\mathrm{init}}$
\FOR{$n=1$ to $N$}
\STATE sample $\mathcal{B}=\{\xi_i\}_{i=1}^{B}$ with
$\tau_i\sim p(\tau)$ and $\xi_i\sim\mathcal{D}_{\tau_i}$
\FOR{$r=1$ to $R$}
\STATE $\tilde I_i\leftarrow
\Omega(\theta;I_i,C_i,P_i,M_i,\psi^{\star})$,
$i=1,\ldots,B$
\STATE $\mathcal{L}\leftarrow\frac{1}{B}\sum_i
\mathcal{L}_{\mathrm{tgt}}
(\tilde I_i,L_i;\mathbf{a}^{\star},\mathcal{J})$
\STATE $\theta\leftarrow
\mathrm{AdamW}(\theta,\nabla_{\theta}\mathcal{L})$
\STATE $\theta\leftarrow\mathrm{clip}(\theta,0,1)$
\ENDFOR
\ENDFOR
\RETURN texture $\theta^{\star}$
\end{algorithmic}
\end{algorithm}

\section{Experiments}

\subsection{Experimental Setup}

\subsubsection{Victim Policies and Task Suites.}

We attack two VLA models that instantiate the action interfaces described above. For the autoregressive action-token interface, we use the suite-specific OpenVLA~\citep{kim2024openvla} checkpoints \texttt{openvla-7b-finetuned-libero-spatial} and \texttt{openvla-7b-finetuned-libero-goal}. For the flow-matching action-chunk interface, we use the official $\pi_{0.5}$~\citep{intelligence2025pi05visionlanguageactionmodelopenworld} checkpoint \texttt{pi05\_libero}. OpenVLA consumes the agent view, whereas $\pi_{0.5}$ consumes both the agent and wrist views together with proprioceptive state; we composite the rendered textured object into every visual stream while leaving proprioception unchanged. We evaluate both policies on LIBERO-Spatial and LIBERO-Goal~\citep{liu2023libero}, each comprising $10$ language-conditioned manipulation tasks. Both suites contain the same plate and bowl objects across all tasks, enabling within-suite cross-task evaluation and controlled cross-suite transfer without changing the target object. Because OpenVLA uses suite-specific fine-tuned checkpoints, its cross-suite evaluation transfers the texture across both task suites and fine-tuned checkpoints, whereas $\pi_{0.5}$ retains the same checkpoint across suites.

\begin{table*}[t]
\centering
\renewcommand{\arraystretch}{0.87}
\small
\setlength{\tabcolsep}{17.5pt}
\newcommand{\tdsneg}[1]{\llap{$-\,$}#1}
\begin{tabular}{lcccccccc}
\toprule
\multirow{3}{*}{Condition} & \multicolumn{4}{c}{OpenVLA} & \multicolumn{4}{c}{$\pi_{0.5}$}\\
\cmidrule(lr){2-5}\cmidrule(lr){6-9}
& \multicolumn{2}{c}{Spatial} & \multicolumn{2}{c}{Goal} & \multicolumn{2}{c}{Spatial} & \multicolumn{2}{c}{Goal}\\
\cmidrule(lr){2-3}\cmidrule(lr){4-5}\cmidrule(lr){6-7}\cmidrule(lr){8-9}
 & plate & bowl & plate & bowl & plate & bowl & plate & bowl\\
\midrule
\multicolumn{9}{l}{SR (\%, lower indicates stronger task disruption)}\\
\quad Clean         & \multicolumn{2}{c}{84.0} & \multicolumn{2}{c}{80.0} & \multicolumn{2}{c}{99.0} & \multicolumn{2}{c}{97.0}\\
\quad Original      & 80.0 & 77.0 & 77.0 & 78.0 & 100.0 & 99.0 & 96.0 & 95.0\\
% \quad All-black     & 65 & 47 & -- & -- & 97 & 94 & 94 & 95\\
\quad Gaussian      & 80.0 & 53.0 & 67.0 & 62.0 & 97.0 & 98.0 & 97.0 & 95.0\\
\quad UniTexture    & \textbf{53.0} & \textbf{25.0} & \textbf{58.0} & \textbf{29.0} & \textbf{33.0} & \textbf{40.0} & \textbf{77.0} & \textbf{72.0}\\
\midrule
\multicolumn{9}{l}{TDS / TDA ( TDS for rendered conditions; TDA for Clean)}\\
\quad Clean         & \multicolumn{2}{c}{$-$19.2} & \multicolumn{2}{c}{$-$23.5} & \multicolumn{2}{c}{$-$21.0} & \multicolumn{2}{c}{$-$26.9}\\
\quad Original      & 0.2 & \tdsneg{0.8} & \tdsneg{0.1} & 0.4 & \tdsneg{0.1} & 0.2 & 0.1 & 0.3\\
% \quad All-black     & 3.1 & 1.9 & -- & -- & 3.5 & 4.2 & 1.6 & 1.6\\
\quad Gaussian      & 1.5 & 0.6 & 1.6 & 2.2 & 1.8 & 0.8 & 0.4 & 0.3\\
\quad UniTexture    & \textbf{3.0} & \tdsneg{2.4} & \textbf{1.9} & \textbf{2.5} & \textbf{10.6} & \textbf{17.9} & \textbf{6.6} & \textbf{4.9}\\
\midrule
\multicolumn{9}{l}{pDHR / DHR (\%; pDHR for rendered conditions; DHR for Clean)}\\
\quad Clean         & \multicolumn{2}{c}{14.8} & \multicolumn{2}{c}{14.2} & \multicolumn{2}{c}{14.0} & \multicolumn{2}{c}{16.0}\\
\quad Original      & 23.4 & 29.1 & 21.8 & 26.0 & 16.6 & 39.3 & 23.0 & 28.5\\
% \quad All-black     & 30.9 & 33.5 & -- & -- & 54.5 & 57.6 & 37.0 & 39.1\\
\quad Gaussian      & 27.4 & 30.3 & 27.6 & 28.0 & 44.0 & 42.6 & 31.7 & 29.6\\
\quad UniTexture    & \textbf{37.6} & \textbf{33.3} & \textbf{33.8} & \textbf{32.5} & \textbf{61.0} & \textbf{65.5} & \textbf{49.3} & \textbf{44.7}\\
\bottomrule
\end{tabular}
\caption{Within-suite results and non-adversarial controls. Each UniTexture is jointly optimized across all ten tasks of one suite and deployed without task-specific refinement; each model-object-suite setting is evaluated over $100$ episodes. Clean observations bypass rendering, so each clean result spans the corresponding plate and bowl columns. }
\label{tab:main}
\end{table*}

\subsubsection{Attack Configuration.}

For each policy-suite pair, we optimize separate object-bound textures for the \emph{plate} and \emph{bowl}, with each texture shared across all $10$ tasks in the suite. Unless otherwise stated, we target the $z$-translation dimension ($j=2$) with $a^\star=+1$, corresponding to an upward command. %; other target dimensions and directions are examined in the ablation study.
Each texture is optimized for $1001$ outer iterations using AdamW with batch size $8$, an initial learning rate of $10^{-3}$, and cosine decay. For each sampled minibatch, we perform $50$ consecutive texture updates before sampling the next minibatch. We render at $224\times224$ using PyTorch3D~\citep{ravi2020accelerating} with soft Phong shading and $8$ faces per pixel.

\subsubsection{Evaluation Protocol.}
Each setting covers all $10$ tasks with $10$ rollouts each, totaling $100$ episodes. Following the official LIBERO initialization and OpenVLA evaluation protocols~\citep{liu2023libero,kim2024openvla}, we execute $10$ dummy actions after resetting the simulator and begin evaluation only after objects settle. OpenVLA predicts one action per step, whereas $\pi_{0.5}$ executes the first five actions of each predicted chunk before re-querying, following the OpenPI LIBERO protocol. At each query, clean and attacked predictions are computed from the same simulator state and, for $\pi_{0.5}$, the same flow-sampling noise. Only attacked actions are executed; clean predictions serve as step-aligned counterfactuals along the attacked trajectory.

Our metrics distinguish nominal behavior, attack-induced action shifts, and task outcomes. For an episode with $T$ steps, let $i$ index the step, $j$ the targeted action dimension, and $\sigma$ the attacker-specified direction.

We first characterize the absolute clean behavior using the \emph{Target Direction Action}(TDA) and \emph{Direction Hit Rate}(DHR):
\[
\begin{aligned}
\mathrm{TDA}
&=
\frac{1}{T}\sum_{i=1}^{T}
\sigma a^{\mathrm{clean}}_{i,j} \times 100,
\\
\mathrm{DHR}
&=
\frac{1}{T}
\sum_{i=1}^{T}
\mathbf{1}\left\{
\sigma a^{\mathrm{clean}}_{i,j}
\geq 0.01
\right\}
\times 100\%.
\end{aligned}
\]
Here, $\mathbf{1}$ is the indicator function, returning $1$ when the condition holds and $0$ otherwise. TDA measures the mean signed clean action along the target direction, while DHR is the percentage of clean steps following that direction with magnitude at least $0.01$.

For rendered conditions, we measure the change induced by the texture relative to the paired clean prediction. The \emph{Target Direction Shift}(TDS) and \emph{Paired Direction Hit Rate}(pDHR) are
\[
\begin{aligned}
\mathrm{TDS}
&=
\frac{1}{T}\sum_{i=1}^{T}
\sigma
\left(
a^{\mathrm{adv}}_{i,j}
-
a^{\mathrm{clean}}_{i,j}
\right) \times 100,
\\
\mathrm{pDHR}
&=
\frac{1}{T}\sum_{i=1}^{T}
\mathbf{1}\left\{
\sigma
\left(
a^{\mathrm{adv}}_{i,j}
-
a^{\mathrm{clean}}_{i,j}
\right)
\geq 0.01
\right\} \times 100\%.
\end{aligned}
\]
TDA and DHR characterize the clean policy's nominal directional behavior, whereas TDS and pDHR isolate the directional effect attributable to the attack. A positive TDS therefore provides direct evidence that the attack steers the predicted action toward the attacker-specified direction relative to a state-matched clean prediction.

Finally, \emph{Task Success Rate}(SR) is the percentage of successfully completed episodes, with lower values indicating stronger task disruption.

\subsubsection{Non-Adversarial Texture Controls.}
To disentangle targeted optimization from renderer-induced discrepancies and appearance changes, we include a renderer-free clean reference and two non-adversarial texture controls. \emph{Clean observation} uses the observation image produced directly by LIBERO, bypassing our rendering and compositing pipeline. \emph{Rendered original texture} passes the object's original asset texture through the frozen calibrated renderer and the same compositing procedure used by UniTexture, isolating changes introduced by rendering and composition. \emph{Rendered Gaussian-noise texture} instead uses an unoptimized Gaussian-noise texture while keeping the rendering and compositing procedure unchanged, sampled once from $\mathcal{N}(0.5,0.2^2)$ and held fixed throughout evaluation. Neither texture control is optimized using $\mathcal{L}_{\mathrm{tgt}}$. All three conditions use the same tasks, initial states, rollout budgets, and policy execution settings as the adversarial evaluation.

\begin{figure*}[!t]
\centering
\includegraphics[width=0.925\textwidth]{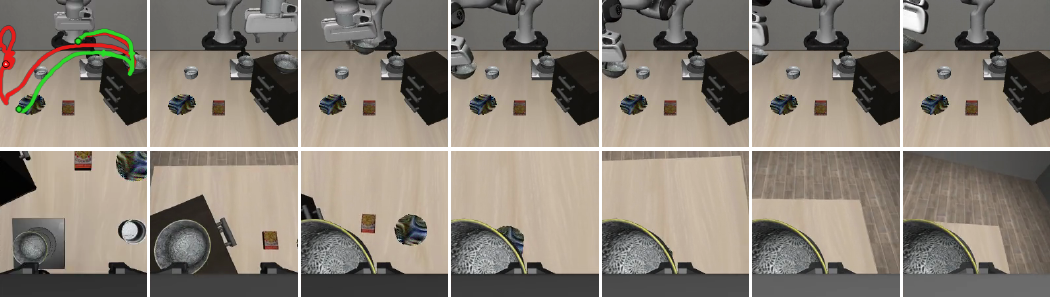}
\caption{Representative failed $\pi_{0.5}$ rollout under a UniTexture attack on the plate in LIBERO-Spatial. The leftmost agent view shows projected gripper-center trajectories for the clean rollout (green; TDA $=-12.4$, DHR $=12.5\%$) and attacked rollout (red; TDS $=18.3$, pDHR $=58.6\%$), interpolated from centers tracked every $20$ frames. The remaining panels show the attacked rollout in temporal order, with agent views above wrist views.}
\label{fig:episode91_rollout}
\end{figure*}

\subsection{Results}

\subsubsection{Within-Suite Cross-Task Attack.}
Table~\ref{tab:main} shows that a single texture jointly optimized across a suite reduces task success in all eight model--object--suite settings without task-specific refinement. For OpenVLA, SR drops from the renderer-free clean baselines of $84\%$ on Spatial and $80\%$ on Goal to $53\%/25\%$ (plate/bowl) and $58\%/29\%$, respectively; for $\pi_{0.5}$, it falls from $99\%$ and $97\%$ to $33\%/40\%$ and $77\%/72\%$. UniTexture also yields lower SR than both the rendered original-texture and Gaussian-noise controls in every setting, indicating that the degradation is not explained by rendering and compositing discrepancies or an arbitrary texture change alone. At the suite-aggregate level, the reduction relative to clean ranges from $20$ points for $\pi_{0.5}$ on Goal-plate to $66$ points on Spatial-plate, showing broad but architecture- and suite-dependent effectiveness.
%% 以上已经确定

%A positive TDS does not necessarily imply a complete reversal of the nominal action. For Spatial--bowl, the clean TDA is $-20$ in unscaled action units, indicating an average downward motion, while UniTexture induces a target-aligned shift of $+0.176$. The resulting attacked action therefore remains slightly negative on average ($-0.024$), but the attack cancels approximately $88\%$ of the nominal downward tendency. It consequently delays the descent rather than reversing it throughout the rollout, although upward reversals occur at individual time steps. Such sustained attenuation can still delay contact with the object and disrupt task completion.

\begin{figure}[t]
\centering
\includegraphics[width=\columnwidth]{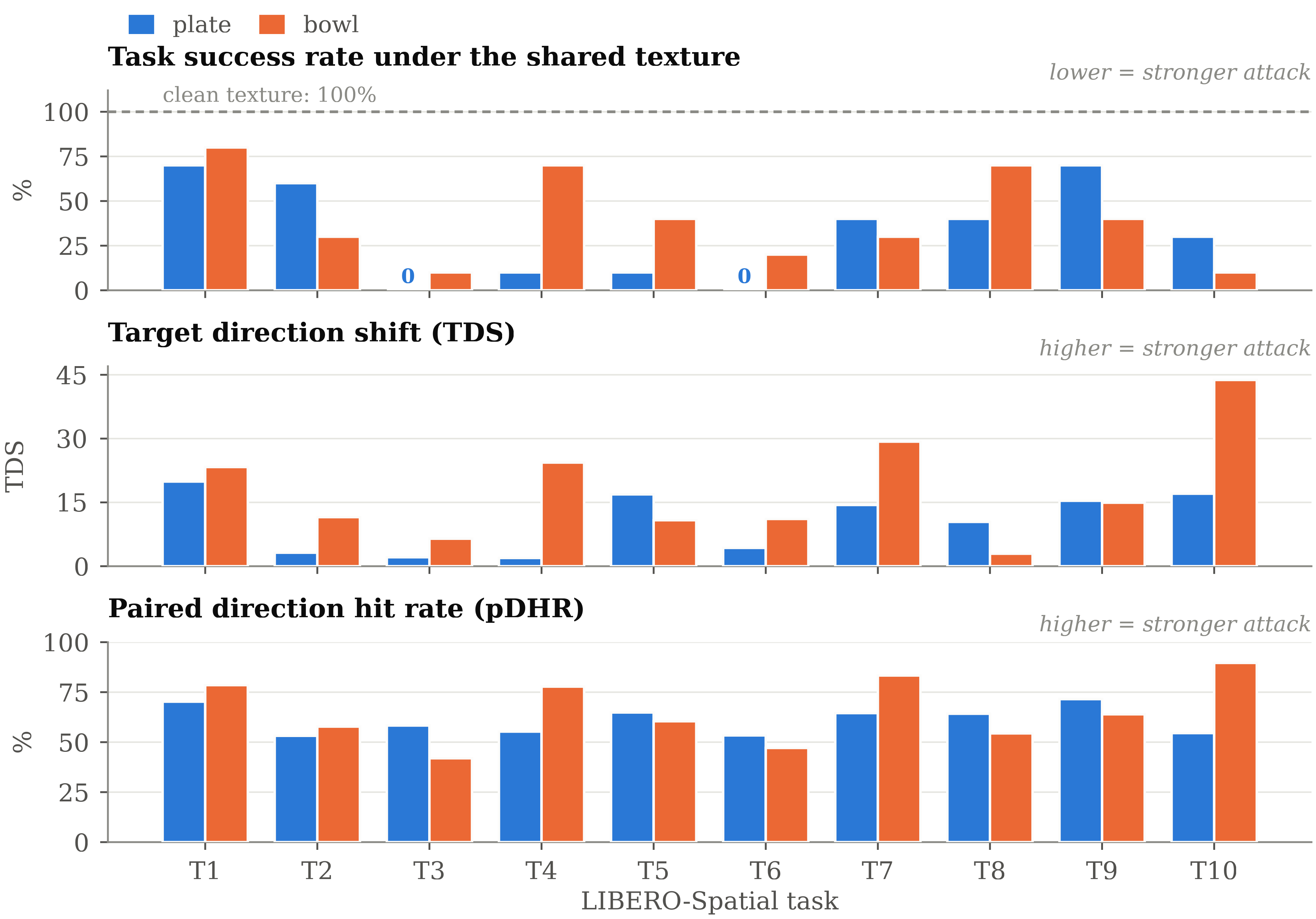}
\caption{Per-task results of the two object-specific UniTexture attacks against $\pi_{0.5}$ on LIBERO-Spatial. The plate and bowl textures yield aggregate SRs of $33\%$ and $40\%$, respectively, in Table~\ref{tab:main}. Each texture is jointly optimized across all ten tasks and deployed unchanged, with $10$ rollouts per task. T1--T10 follow the suite's standard task order; the panels show SR, TDS, and pDHR from top to bottom, and the dashed line marks the $100\%$ clean-reference SR. Both textures produce positive TDS on every task, while the resulting reduction in SR varies across tasks.}
\label{fig:pertask}
\end{figure}
%% 以上正在写

\subsubsection{Targeted Control versus Disruption.}
For $\pi_{0.5}$, UniTexture achieves both targeted control and task disruption, most clearly on Spatial. The TDS reaches $10.6/17.9$, and pDHR reaches $61.0\%/65.5\%$ for plate/bowl, while SR falls to $33\%/40\%$. On Goal, positive TDS of $6.6/4.9$ and pDHR of $49.3\%/44.7\%$ coexist with substantially higher SR of $77\%/72\%$. Thus, a persistent target-aligned shift does not necessarily translate into task failure.

OpenVLA exhibits a different pattern. UniTexture reduces SR to $53\%/25\%$ on Spatial and $58\%/29\%$ on Goal, but its TDS remains modest and becomes negative ($-2.4$) in the Spatial-bowl setting. This setting produces OpenVLA's largest success reduction despite shifting the action in the opposite direction on average, directly separating disruption from targeted control. OpenVLA also has lower clean SR than $\pi_{0.5}$ on the evaluated suites ($80$--$84\%$ versus $97$--$99\%$), which should temper cross-policy comparisons of absolute attacked SR.
We attribute this decoupling to OpenVLA's discrete autoregressive action interface. The token-level objective can alter the targeted action token and propagate its effect through subsequent decoding, disrupting the action sequence without inducing a coherent signed displacement in the targeted dimension. By contrast, the flow-matching objective directly steers a continuous action chunk, yielding the larger positive TDS and pDHR observed for $\pi_{0.5}$.

Gaussian-noise textures already produce nontrivial pDHR, reaching $44.0\%$, so pDHR should be interpreted relative to its matched control. Nevertheless, UniTexture exceeds the corresponding Gaussian pDHR in all eight settings. Overall, UniTexture provides stronger target-aligned control for $\pi_{0.5}$, whereas OpenVLA can be substantially disrupted without reliable steering in the attacker-specified direction.

Figure~\ref{fig:episode91_rollout} illustrates how target-aligned action shifts accumulate into task-level disruption. In this episode, the clean behavior predominantly opposes the targeted $+z$ direction, with a TDA of $-12.4$ and a DHR of $12.53\%$. UniTexture nevertheless induces a large positive paired shift, reaching a TDS of $18.3$ and a pDHR of $58.6\%$. Compared with the smooth clean reference trajectory (green), the attacked trajectory (red) alternates between downward corrective motions and attack-induced upward motions before task failure.

\subsubsection{Task-Level Coverage of Shared Textures.}
Figure~\ref{fig:pertask} expands the $\pi_{0.5}$ LIBERO-Spatial results in Table~\ref{tab:main} into their task-level outcomes. The targeted effect spans the entire suite: both object-specific shared textures produce positive TDS on all ten tasks, with pDHR ranging from $42\%$ to $90\%$ across all $20$ task--object pairs. The attack also reduces SR below the $100\%$ per-task reference in every pair, including complete failure on two plate tasks. These results show that the suite-level effectiveness is not driven by a small subset of vulnerable tasks: each shared texture induces the specified action direction throughout the task suite without task-specific refinement. Differences in residual SR instead characterize how individual tasks respond to the induced action shift, rather than whether the attack reaches them.

\begin{table}[!t]
\centering
\renewcommand{\arraystretch}{0.87}
\small
\setlength{\tabcolsep}{6.5pt}
\newcommand{\trtds}[1]{\phantom{1}#1}
\newcommand{\trtdsneg}[1]{\phantom{1}\llap{$-\,$}#1}
\begin{tabular}{llcccc}
\toprule
Policy & Suite & Object & SR & TDS & pDHR (\%)\\
\midrule
\multicolumn{6}{l}{\emph{Cross-suite} (same policy, texture reused on the other suite)}\\
OpenVLA & S$\rightarrow$G & plate & 65.0 & \trtds{1.5} & 33.0\\
OpenVLA & S$\rightarrow$G & bowl  & 39.0 & \trtds{5.3} & 35.2\\
OpenVLA & G$\rightarrow$S & plate & 72.0 & \trtds{2.9} & 32.5\\
OpenVLA & G$\rightarrow$S & bowl  & 24.0 & \trtdsneg{1.6} & 31.9\\
$\pi_{0.5}$ & S$\rightarrow$G & plate & 74.0 & \trtds{3.0} & 42.7\\
$\pi_{0.5}$ & S$\rightarrow$G & bowl  & 81.0 & \trtds{5.1} & 44.2\\
$\pi_{0.5}$ & G$\rightarrow$S & plate & 78.0 & \trtds{6.6} & 60.3\\
$\pi_{0.5}$ & G$\rightarrow$S & bowl  & 53.0 & 13.4 & 59.5\\
\midrule
\multicolumn{6}{l}{\emph{Cross-model} (same suite, texture reused on the other policy)}\\
O$\rightarrow\pi$ & Spatial & plate & 92.0 & \trtds{3.4} & 52.2\\
O$\rightarrow\pi$ & Spatial & bowl  & 95.0 & \trtds{2.2} & 50.0\\
O$\rightarrow\pi$ & Goal    & plate & 94.0 & \trtds{1.0} & 36.6\\
O$\rightarrow\pi$ & Goal    & bowl  & 97.0 & \trtds{1.0} & 36.3\\
$\pi\rightarrow$O & Spatial & plate & 61.0 & \trtds{3.4} & 33.9\\
$\pi\rightarrow$O & Spatial & bowl  & 26.0 & \trtds{0.7} & 34.8\\
$\pi\rightarrow$O & Goal    & plate & 55.0 & \trtds{2.5} & 32.6\\
$\pi\rightarrow$O & Goal    & bowl  & 41.0 & \trtds{1.7} & 30.7\\
\bottomrule
\end{tabular}
\caption{Cross-suite and cross-model transfer without re-optimization. Arrows indicate the optimization-to-evaluation direction; S/G and O/$\pi$ denote Spatial/Goal and OpenVLA/$\pi_{0.5}$, respectively. OpenVLA cross-suite transfer also changes the suite-specific checkpoint.}
\label{tab:transfer}
\end{table}

\subsubsection{Cross-Suite and Cross-Model Transfer.}
Table~\ref{tab:transfer} evaluates direct texture reuse when either the task suite or the victim policy changes. Cross-suite transfer reduces SR below both the renderer-free clean baseline and the corresponding Gaussian-texture control in all eight settings. For $\pi_{0.5}$, the directional objective also transfers consistently: TDS remains positive in all four settings, with pDHR ranging from $42.7\%$ to $60.3\%$. Cross-model transfer is asymmetric in task disruption. Textures optimized for $\pi_{0.5}$ reduce OpenVLA SR to $26\%$--$61\%$, whereas textures optimized for OpenVLA leave $\pi_{0.5}$ SR at $92\%$--$97\%$.

This asymmetry reflects a substantial difference in robustness to object appearance. Under the matched non-adversarial controls, replacing the rendered original texture with a Gaussian-noise texture reduces OpenVLA's Spatial--bowl SR from $77\%$ to $53\%$, whereas $\pi_{0.5}$ changes only from $99\%$ to $98\%$. This sensitivity becomes more pronounced under adversarial textures. The directly optimized OpenVLA--Spatial--bowl texture, the transferred OpenVLA--Goal--bowl texture, and the transferred $\pi_{0.5}$--Spatial--bowl texture produce nearly identical SRs of $25\%$, $24\%$, and $26\%$, respectively. Despite this severe disruption, all three exhibit near-zero or negative TDS and pDHR of only $31.9\%$--$34.8\%$, close to the Gaussian control of $30.3\%$. Weak directional metrics therefore do not indicate an ineffective attack: the textures strongly disrupt OpenVLA, but do so by destabilizing its visual--action mapping rather than by inducing a coherent shift in the specified direction. We attribute this failure mode to OpenVLA's limited robustness to object appearance, which allows texture-induced instability to overwhelm the directional signal encouraged by the targeted objective. In contrast, $\pi_{0.5}$ remains stable under the Gaussian texture and exhibits strong targeted control in the same Spatial--bowl setting, reaching TDS of $17.9$ and pDHR of $65.5\%$.

\begin{table}[!t]
\centering
\renewcommand{\arraystretch}{0.87}
\small
\setlength{\tabcolsep}{11pt}
\begin{tabular}{cccccc}
\toprule
Dim.  & Semantics & SR & TDS & pDHR (\%)\\
\midrule
0 & $+x$ & 62.0 & 4.8 & 48.1\\
0 & $-x$ & 38.0 & 5.6 & 55.3\\
1 & $+y$ & 68.0 & 3.8 & 50.8\\
1 & $-y$ & 45.0 & 4.3 & 55.1\\
2 & $+z$ & \textbf{33.0} & \textbf{10.6} & \textbf{61.0}\\
2 & $-z$ & 91.0 & 2.7 & 36.4\\
\bottomrule
\end{tabular}
\caption{Effect of the targeted action dimension and direction ($\pi_{0.5}$-Spatial-plate). The upward $z$ target used throughout the paper is the most effective choice, while its downward counterpart is the least effective.}
\label{tab:ablation}
\end{table}

\subsubsection{Which Action Target Matters.}
Table~\ref{tab:ablation} varies the targeted action dimension and direction while holding the policy, object, task suite, and optimization configuration fixed. Under our evaluation coordinate convention, $+z$ and $-z$ denote upward and downward motion, $+y$ and $-y$ denote leftward and rightward motion in the camera view, and $+x$ and $-x$ denote motion toward and away from the camera, respectively.

The evaluated horizontal targets remain controllable, with pDHR between $48.1\%$ and $55.1\%$, but produce weaker task disruption, with SR between $45\%$ and $68\%$. The opposite vertical target, $-z$, is the least effective on both criteria, retaining an SR of $91\%$ while reaching only $2.7$ TDS and $36.4\%$ pDHR. We attribute this directional asymmetry to the manipulation geometry: a persistent upward bias pulls the end effector away from the interaction surface and disrupts reaching and contact, whereas a downward bias often agrees with the nominal approach motion.

\section{Conclusion}
UniTexture shows that task diversity does not inherently protect a multitask VLA from a shared visual attack surface: one object-bound texture can induce cross-task adversarial effects without per-task refinement. By jointly optimizing the texture over a task distribution, UniTexture turns a persistent object appearance into an attack shared across different instructions, scenes, and trajectories. Results on OpenVLA and $\pi_{0.5}$ demonstrate attacker-specified directional shifts and task disruption, while cross-suite and asymmetric cross-model transfer show that some effects extend beyond the optimization setting. These findings argue that VLA robustness should be evaluated against persistent perturbations shared across tasks, rather than only task-specific attacks.
% \subsubsection{Limitations.}
% Our evaluation covers two VLA policies and two LIBERO task suites, rather than the full diversity of VLA systems and manipulation benchmarks. UniTexture studies cross-task universality for a fixed victim policy and object geometry; cross-suite and cross-model experiments evaluate transfer rather than joint universality across all settings.

\bibliography{ref}

\end{document}